\documentclass{article}

\newif\ifarxiv
\arxivtrue

\usepackage{times}
\usepackage{graphicx} 
\usepackage{subfigure} 
\usepackage[utf8]{inputenc}

\graphicspath{{figures/}}

\usepackage{xfrac} 

\usepackage{natbib}

\usepackage{algorithm}
\usepackage{algorithmic}

\usepackage{amsmath}

\usepackage{url}

\usepackage[accepted]{arxiv}

\def\H{{\mathrm{\bf H}}}
\def\KL{{\mathrm{\bf KL}}}

\def\ss{\vspace*{-.5mm}}
\def\sa{\vspace*{-1mm}}
\def\sb{\vspace*{-2mm}}
\def\sc{\vspace*{-3mm}}
\def\sd{\vspace*{-4mm}}
\def\se{\vspace*{-5mm}}

\icmltitlerunning{Towards Biologically Plausible Deep Learning}

\begin{document} 

\twocolumn[
\icmltitle{Towards Biologically Plausible Deep Learning}

\icmlauthor{Yoshua Bengio$^1$, Dong-Hyun Lee, Jorg Bornschein, Thomas Mesnard and Zhouhan Lin\\}{}
\icmladdress{Montreal Institute for Learning Algorithms, University of Montreal, Montreal, QC, H3C 3J7\\
$^1$CIFAR Senior Fellow}

\icmlkeywords{deep learning, unsupervised learning, auto-encoders, biological learning}

\vskip 0.3in
]

\begin{abstract}

Neuroscientists have long criticised deep learning algorithms as incompatible with current knowledge of neurobiology. We explore more biologically plausible versions of deep representation learning, focusing here mostly on unsupervised learning but developing a learning mechanism that could account for supervised, unsupervised and reinforcement learning. The starting point is that the basic learning rule believed to govern synaptic weight updates (Spike-Timing-Dependent Plasticity) arises out of a simple update rule that makes a lot of sense from a machine learning point of view and can be interpreted as gradient descent on some objective function so long as the neuronal dynamics push firing rates towards better values of the objective function (be it supervised, unsupervised, or reward-driven).  The second main idea is that this corresponds to a form of the variational EM algorithm, i.e., with approximate rather than exact posteriors, implemented by neural dynamics.  Another contribution of this paper is that the gradients required for updating the hidden states in the above variational interpretation can be estimated using an approximation that only requires propagating activations forward and backward, with pairs of layers learning to form a denoising auto-encoder. Finally, we extend the theory about the probabilistic interpretation of auto-encoders to justify improved sampling schemes based on the generative interpretation of denoising auto-encoders, and we validate all these ideas on generative learning tasks.
\sb
\end{abstract}

\sd

\section{Introduction}
\sb

Deep learning and artificial neural networks have taken their
inspiration from brains, but mostly for the form of the computation
performed (with much of the biology, such as the presence of spikes 
remaining to be accounted for). However, what is lacking currently is
a credible {\em machine learning interpretation of the
learning rules} that seem to exist in biological neurons that
would explain {\em efficient} joint training of a deep neural network, i.e., accounting
for {\em credit assignment through a long chain of neural connections}. 
Solving the credit assignment problem therefore means identifying neurons and weights 
that are responsible for a desired outcome and changing parameters accordingly. 
Whereas back-propagation offers a machine learning answer, it is
not biologically plausible, as discussed in the next paragraph. 
Finding a biologically plausible machine
learning approach for credit assignment in deep networks is the
main long-term question to which this paper contributes.

Let us first consider the claim that state-of-the-art deep learning
algorithms rely on mechanisms that seem biologically implausible, such as
gradient back-propagation, i.e., the mechanism for computing the gradient
of an objective function with respect to neural activations and parameters.
The following difficulties can be raised regarding the biological
plausibility of back-propagation: (1) the back-propagation computation
(coming down from the output layer to lower hidden layers) is purely
linear, whereas biological neurons interleave linear and non-linear
operations, (2) if the feedback paths known to exist in the brain (with
their own synapses and maybe their own neurons) were used
to propagate credit assignment by backprop, they would need precise
knowledge of the derivatives of the non-linearities at the operating point
used in the corresponding feedforward computation on the feedforward path\footnote{and
with neurons not all being exactly the same, it could be difficult to match the
right estimated derivatives}, (3) similarly, these
feedback paths would have to use exact symmetric weights (with the same
connectivity, transposed) of the feedforward connections,\footnote{this is
  known as the {\em weight transport}
  problem~\citep{Lillicrap-et-al-arxiv2014}} (4) real neurons communicate
by (possibly stochastic) binary values (spikes), not by clean continuous
values, (5) the computation would have to be precisely clocked to alternate
between feedforward and back-propagation phases (since the latter needs the former's results), 
and (6) it is not clear
where the output targets would come from. The approach proposed in this
paper has the ambition to address all these issues, although some question
marks as to a possible biological implementations remain, and of course
many more detailed elements of the biology that need to be accounted for are not covered
here.

Note that back-propagation is used not just for classical supervised
learning but also for many unsupervised learning algorithms, including all
kinds of auto-encoders: sparse auto-encoders~\citep{ranzato-07-small,Goodfellow2009-short}, denoising
auto-encoders~\citep{VincentPLarochelleH2008-small}, contractive auto-encoders~\citep{Rifai+al-2011-small}, 
and more
recently, variational auto-encoders~\citep{Kingma+Welling-ICLR2014}. Other unsupervised learning
algorithms exist which do not rely on back-propagation, such as the various
Boltzmann machine learning algorithms~\citep{Hinton-bo86,Smolensky86,Hinton06,Salakhutdinov2009-small}.  
Boltzmann machines are
probably the most biologically plausible learning algorithms for deep
architectures that we currently know, but they also face several question
marks in this regard, such as the weight transport problem ((3) above) to
achieve symmetric weights, and the positive-phase vs negative-phase
synchronization question (similar to (5) above).

Our starting point (Sec.~\ref{sec:stdp}) proposes an interpretation of the
main learning rule observed in biological synapses: Spike-Timing-Dependent
Plasticity (STDP). Inspired by earlier ideas~\citep{Xie+Seung-NIPS1999,Hinton-DL2007}, 
we first show via both an intuitive argument and a simulation that STDP could be seen as
stochastic gradient descent if only the neuron was driven by a feedback
signal that either increases or decreases the neuron's firing rate in
proportion to the gradient of an objective function with respect to the
neuron's voltage potential. 

In Sec.~\ref{sec:v-EM} we present the first machine learning interpretation
of STDP that gives rise to efficient credit assignment through multiple layers.
We first argue that the above
interpretation of STDP suggests that neural dynamics (which creates the above
changes in neuronal activations thanks to feedback and lateral connections)
correspond to {\em inference} towards neural configurations that are more
consistent with each other and with the observations (inputs, targets, or
rewards). This view is analogous to the interpretation of inference
in Boltzmann machines while avoiding the need to obtain representative samples
from the stationary distribution of an MCMC. Going beyond Hinton's proposal, 
it naturally suggests that the training procedure corresponds
to a form of variational EM~\citep{emview} (see Sec.\ref{sec:v-EM}), possibly based on MAP (maximum a posteriori)
or MCMC (Markov Chain Monte-Carlo)
approximations. In Sec.~\ref{sec:deep} we show how this mathematical
framework suggests a training procedure for a deep directed generative network 
with many layers of latent variables.
However, the above interpretation would still require to compute some
gradients. Another contribution (Sec.~\ref{sec:bf-gradient}) is to show
that one can estimate these gradients via an approximation that only
involves ordinary neural computation and no explicit derivatives,
following previous (unpublished) work on target propagation~\citep{Bengio-arxiv2014,Lee-et-al-arxiv2014-small}.
We introduce a novel justification for difference target propagation~\citep{Lee-et-al-arxiv2014-small},
exploiting the fact that the proposed learning mechanism can be interpreted as training a denoising auto-encoder.
As discussed in Sec.~\ref{sec:DAE-generator}
these alternative interpretations of the model provide different
ways to sample from it, and we found that better samples could
be obtained.

\sc
\section{STDP as Stochastic Gradient Descent}
\label{sec:stdp}
\sb

Spike-Timing-Dependent Plasticity or STDP is believed to be the main form 
of synaptic change in neurons~\citep{Markram+Sakmann-1995,Gerstner-et-al-1996} 
and it relates the expected
change in synaptic weights to the timing difference between post-synaptic
spikes and pre-synaptic spikes. Although it is the result of experimental
observations in biological neurons, its interpretation as part of a learning
procedure that could explain learning in deep networks remains unclear.
\citet{Xie+Seung-NIPS1999} nicely showed how STDP can correspond
to a differential anti-Hebbian plasticity, i.e., the synaptic change is
proportional the product of pre-synaptic activity and the temporal derivative
of the post-synaptic activity. The question is how this could make
sense from a machine learning point of view.
This paper aims at proposing such an interpretation, starting from
the general idea introduced by~\citet{Hinton-DL2007}, anchoring it
in a novel machine learning interpretation, and extending it
to deep unsupervised generative modeling of the data.

What has been observed in STDP is that the weights change if there is a
pre-synaptic spike in the temporal vicinity of a post-synaptic spike: 
that change is positive if the post-synaptic
spike happens just after the pre-synaptic spike, negative if it happens
just before. Furthermore, the amount of change decays to zero as the
temporal difference between the two spikes increases in magnitude.
We are thus interested in this temporal window around a pre-synaptic
spike during which we could have a post-synaptic spike, before or
after the pre-synaptic spike.

We propose a novel explanation for the STDP curve as a consequence of
an actual update equation which makes a lot of sense from a machine
learning perspective:
\sa
\begin{equation}
 \Delta W_{ij} \propto S_i \dot{V}_j,
\label{eq:delta-w-stdp}
\sa
\end{equation}
where $\dot{V}_j$ indicates the temporal derivative of $V_j$, $S_i$ indicates the
pre-synaptic spike (from neuron $i$), and $V_j$ indicates the
post-synaptic voltage potential (of neuron $j$).

To see how the above update rule can give rise to STDP, 
consider the average effect of the rest of the
inputs into neuron $j$, which induces an average temporal change $\dot{V}_j$
of the post-synaptic potential (which we assume approximately constant throughout the
duration of the window around the pre-synaptic firing event), and assume that at time 0 when
$S_i$ spikes, $V_j$ is below the firing threshold. Let us call $\Delta T$
the temporal difference between the post-synaptic spike and the pre-synaptic spike.

\begin{figure}[htp]
\vspace*{-3mm}
\centerline{\includegraphics[width=.5\textwidth]{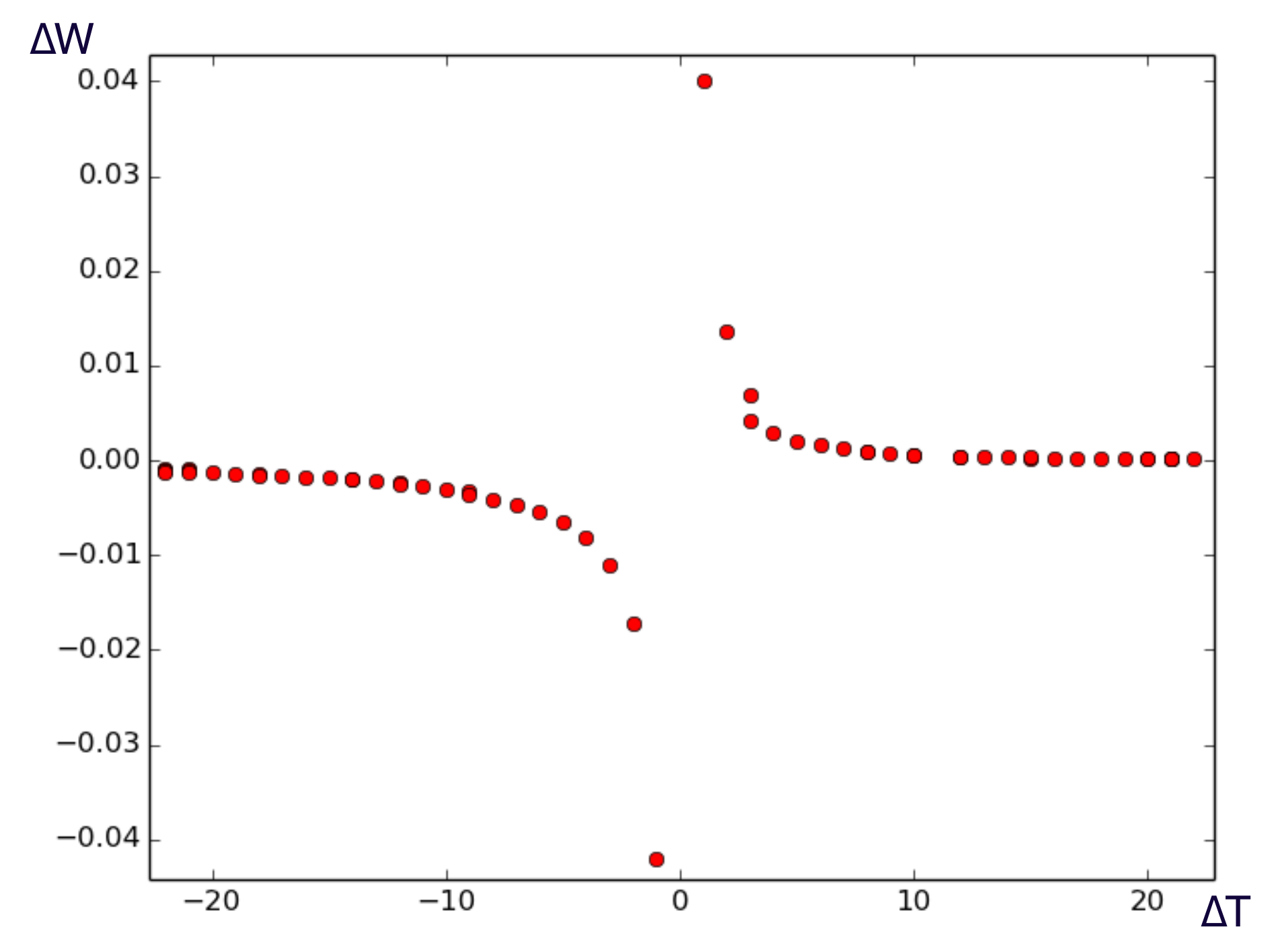}}
\vspace*{-3mm}
\caption{Result of simulation around pre-synaptic spike (time 0) showing indirectly the effect of
a change in the rate of change in the post-synaptic voltage, $\dot{V}_j$ on both
the average time difference between pre- and post-synaptic spikes (horizontal axis, $\Delta T$)
and the average weight change (vertical axis, $\Delta W_{ij}$), when the latter follows
Eq.~\ref{eq:delta-w-stdp}. This corresponds very well to the observed relationship between
$\Delta T$ and $\Delta W_{ij}$ in the biological literature.}
\label{fig:simulated-STDP}
\sc
\end{figure}

First, let us
consider the case where $\dot{V}_j > 0$, i.e., the voltage potential
increases. Depending on the magnitude of $\dot{V}_j$, it will take more or
less time for $V_j$ to reach the firing threshold, i.e. more time for smaller $\dot{V}_j$.
Hence {\em a longer $\Delta T$ corresponds to a smaller $\Delta W_{ij}$}, and
{\em a positive $\Delta T$ to a positive $\Delta W_{ij}$}, as observed for STDP.

Second, let us consider the case where $\dot{V}_j < 0$, i.e., the voltage
potential {\em has been decreasing} (remember that we are considering the average effect
of the rest of the inputs into neuron $j$, and assuming that the temporal slope
of that effect is locally constant). Thus, it is likely that {\em earlier before
the pre-synaptic spike} the post-synaptic voltage $V_j$ had been high enough
to be above the firing threshold. How far back in the past again depends monotonically
on $\dot{V}_j$. Hence {\em a negative $\Delta T$ corresponds to a negative $\Delta W_{ij}$}
and {\em a more negative $\Delta T$ corresponds to a $\Delta W_{ij}$ that is
smaller in magnitude}. This corresponds perfectly to the kind of relationship
that is observed by biologists with STDP (see Figure 7 of \citet{Bi+Poo-1998}
or Figure 1 of~\citet{Sjostrom+Gerstner-2010}, e.g.
at \small{\url{http://www.scholarpedia.org/article/Spike-timing_dependent_plasticity}}). 
In a simulation inspired by the above analysis, we observe
essentially the same curve relating the average $\Delta T$ and the $\Delta W_{ij}$ 
that is associated with it, as illustrated in Figure~\ref{fig:simulated-STDP}.

Clearly, the consequence of Eq.~\ref{eq:delta-w-stdp} 
  is that {\bf if the change $\Delta V_j$
  corresponds to improving some objective function $J$, then STDP
  corresponds to approximate stochastic gradient descent in that objective
  function}. With this view, STDP would implement the delta rule
 (gradient descent on a one-layer network) if the post-synaptic activation
 changes in the direction of the gradient.

\sc
\section{Variational EM with Learned Approximate Inference}
\label{sec:v-EM}
\sb

To take advantage of the above statement, the dynamics of the neural
network must be such that {\em neural activities move towards better values
  of some objective function $J$}. Hence we would like to define such an
objective function in a way that is consistent with the actual neural
computation being performed (for fixed weights $W$), in the sense that the
expected temporal change of the voltage potentials approximately
corresponds to increases in $J$. In this paper, we are going to consider
the voltage potentials as the central variables of interest which influence
$J$ and consider them as latent variables $V$ (denoted $h$ below to keep
machine learning interpretation general), while we will consider the
actual spike trains $S$ as non-linear noisy corruptions of $V$, a form
of quantization (with the ``noise level'' controlled either by the integration
time or the number of redundant neurons in an ensemble~\citep{Legenstein+Maass-2014}). This view
makes the application of the denoising auto-encoder theorems discussed in
Sec.~\ref{sec:DAE-generator} more straightforward.

The main contribution of this paper is to propose and give support to the
hypothesis that $J$ comes out of a {\em variational bound} on the
likelihood of the data.  Variational bounds have been
proposed to justify various learning algorithms for generative
models~\citep{Hinton95} (Sec.~\ref{sec:related-work}). To keep the mapping to biology
open, consider such bounds and the associated criteria
that may be derived from them, using an abstract notation with
{\em observed variable} $x$ and {\em latent variable} $h$.
If we have a model $p(x,h)$ of their joint distribution,
as well as some approximate inference mechanism defining a 
conditional distribution $q^*(H|x)$,
the observed data log-likelihood $\log p(x)$ can be decomposed as
\sa
\begin{align}
\label{eq:ll-decomposition}
 \log p(x) =& \log p(x) \sum_h q^*(h|x) \nonumber \\
  =& \sum_h q^*(h|x) \log \frac{p(x,h)q^*(h|x)}{p(h|x)q^*(h|x)} \nonumber \\
  =& E_{q^*(H|x)}[ \log p(x,H) ] + \H[q^*(H|x)] \nonumber \\
  & \hspace*{15mm} + \KL(q^*(H|x)||p(H|x)),
\sb
\end{align}
where $\H[ ]$ denotes entropy and $\KL( || )$ the Kullback-Leibler (KL) divergence, and
where we have used sums but integrals should be considered when the variables are continuous.
Since both the entropy and the KL-divergence are non-negative, we can either bound the log-likelihood
via
\sa
\begin{equation}
\label{eq:bound1}
  \log p(x) \geq E_{q^*(H|x)}[ \log p(x,H) ] + \H[q^*(H|x)],
\end{equation}
or if we care only about optimizing $p$,
\sa
\begin{equation}
\label{eq:bound2}
  \log p(x) \geq E_{q^*(H|x)}[ \log p(x,H) ].
\sa
\end{equation}
The idea of variational bounds as proxies for the log-likelihood is that {\em as far as 
optimizing $p$ is concerned, i.e., dropping the entropy term which does not depend on $p$, 
the bound becomes tight when $q^*(H|x)=p(H|x)$}. This suggests that $q^*(H|x)$
should approximate $p(H|x)$. Fixing $q^*(H|x)=p(H|x)$ and optimizing $p$ with $q$ fixed is
the EM algorithm. Here (and in general) this is not possible so we consider variational
methods in which $q^*(H|x)$ approximates but does not reach $p(H|x)$.
This variational bound has recently been used to justify another biologically
plausible update rule~\citep{Rezende+Gerstner-2014}, which however relied
on the REINFORCE algorithm~\citep{Williams-1992} 
rather than on inference to obtain credit assignment to internal neurons.

We propose to decompose $q^*(H|x)$ in two components: {\em parametric initialization}
$q_0(H|x)=q(H|x)$ and {\em iterative inference},
implicitly defining $q^*(H|x)=q_T(H|x)$ via
a deterministic or stochastic update, or transition operator
\sa
\begin{equation}
 q_t(H|x) = A(x) \; q_{t-1}(H|x).
\end{equation}
The variational bound suggests that $A(x)$ should gradually bring $q_t(H|x)$
closer to $p(H|x)$. At the same time, to make sure that a few steps will be sufficient
to approach $p(H|x)$, one may add a term in the objective function to make $q_0(H|x)$
closer to $p(H|x)$, as well as to encourage $p(x,h)$ to favor solutions $p(H|x)$
that can be easily approximated by $q_t(H|x)$ even for small $t$. 

For this purpose, consider as training objective a regularized
variational MAP-EM criterion (for a given $x$):
\sa
\begin{equation}
\label{eq:J}
 J = \log p(x,h) + \alpha \log q(h|x),
\end{equation}
where $h$ is a {\em free variable} (for each $x$)
initialized from $q(H|x)$ and then iteratively updated to
approximately maximize $J$. The total objective function is just the average of $J$
over all examples after having performed inference (the approximate maximization over $h$
for each $x$).
A reasonable variant would not just encourage $q=q_0$
to generate $h$ (given $x$), but all the $q_t$'s for $t>0$ as well.
Alternatively, the iterative inference could be performed by
{\em stochastically} increasing $J$, i.e., via a Markov chain which
may correspond to probabilistic inference with spiking neurons~\citep{Pecevski-et-al-2011}.
The corresponding variational MAP or variational MCMC algorithm would be as in
Algorithm~\ref{alg:variational-MAP}. For the stochastic version
one would inject noise when updating $h$.
Variational MCMC~\citep{deFreitas-et-al-UAI2001} can be used to approximate
the posterior, e.g., as in the model from~\citet{Salimans-et-al-arxiv2014}.
However, a rejection step does not look very biologically plausible
(both for the need of returning to a previous state and for the
need to evaluate the joint likelihood, a global quantity). On the
other hand, a {\em biased} MCMC with no rejection step, such as the stochastic gradient
Langevin MCMC of~\citet{Welling+Teh-ICML2011}
can work very well in practice.

\begin{algorithm}[htb]
\ss
\begin{minipage}{0.5\textwidth}
\caption{Variational MAP (or MCMC) SGD algorithm for gradually improving the agreement
between the values of the latent variables $h$ and the observed data $x$.
$q(h|x)$ is a learned parametric initialization for $h$, $p(h)$ is
a parametric prior on the latent variables, and $p(x|h)$ specifies how
to generate $x$ given $h$. Objective function $J$ is defined in Eq.~\ref{eq:J}
Learning rates $\delta$ and $\epsilon$ respectively control the optimization of $h$ and of
parameters $\theta$ (of both $q$ and $p$). 
}
\label{alg:variational-MAP}
\begin{algorithmic}
\STATE Initialize $h \sim q(h|x)$
\FOR {$t=1$ to $T$}
  \STATE $h \leftarrow h + \delta \frac{\partial J}{\partial h}$ \hspace*{3mm} (optional: add noise for MCMC)
\ENDFOR
\STATE $\theta \leftarrow \theta + \epsilon \frac{\partial J}{\partial \theta}$
\end{algorithmic}
\end{minipage}
\ss
\end{algorithm}

\sc
\section{Training a Deep Generative Model}
\label{sec:deep}
\sb

There is strong biological evidence of a distinct pattern of connectivity between cortical
areas that distinguishes between ``feedforward'' and ``feedback'' connections~\citep{Douglas-et-al-1989}
at the level of the microcircuit of cortex (i.e., feedforward and feedback connections do not
land in the same type of cells). Furthermore, the feedforward connections form 
a directed acyclic graph with nodes (areas) updated in a particular order, e.g., in
the visual cortex~\citep{Felleman+VanEssen-1991}. 
So consider Algorithm~\ref{alg:variational-MAP} with
$h$ decomposed into multiple layers, with the conditional
independence structure of a directed graphical model structured as a chain,
both for $p$ (going down) and for $q$ (going up):
\sa
\begin{align}
\label{eq:deep-decomposition}
p(x,h) &= p(x|h^{(1)}) \left( \prod_{k=1}^{M-1} p(h^{(k)}|h^{(k+1)}) \right) p(h^{(M)}) \nonumber \\
q(h|x) &= q(h^{(1)}|x) \prod_{k=1}^{M-1} q(h^{(k+1)}|h^{(k)}). 
\end{align}
This clearly decouples the updates associated with each layer, for both $h$ and $\theta$,
making these updates ``local'' to the layer $k$, based on ``feedback'' from layer $k-1$
and $k+1$. Nonetheless, thanks to the iterative nature of the updates of $h$, all
the layers are interacting via both feedforward ($q(h^{(k)}|h^{(k-1)})$)
and feedback ($p(h^{(k)}|h^{(k+1)})$ paths. Denoting $x=h^{(0)}$ to simplify
notation, the $h$ update would thus consist
in moves of the form
\sa
\begin{align}
\label{eq:h-gradient}
  h^{(k)} \leftarrow & h^{(k)} + \delta \frac{\partial}{\partial h^{(k)}} 
    \left( \log ( p(h^{(k-1)} | h^{(k)}) p(h^{(k)} | h^{(k+1)}) ) \right. \nonumber \\
    & \hspace*{14mm} + \left. \alpha \log ( q(h^{(k)} | h^{(k-1)}) q(h^{(k+1)} | h^{(k)}) ) \right),
\se
\end{align}
where $\alpha$ is as in Eq.~\ref{eq:J}.
No back-propagation is needed for the above derivatives when $h^{(k)}$ is
on the left hand side of the conditional probability bar. Sec.~\ref{sec:bf-gradient}
deals with the right hand side case. 
For the left hand side case, e.g., $p(h^{(k)} | h^{(k+1)})$ a conditional Gaussian with mean $\mu$
and variance $\sigma^2$, the gradient with respect to $h^{(k)}$ is simply
$\frac{\mu - h^{(k)}}{\sigma^2}$.
Note that there is an interesting interpretation of such a deep model: the layers above $h^{(k)}$
provide a complex implicitly defined prior for $p(h^{(k)})$.

\sc
\section{Alternative Interpretations as Denoising Auto-Encoder}
\label{sec:DAE-generator}
\sb

By inspection of Algorithm~\ref{alg:variational-MAP}, one can observe that
this algorithm trains $p(x|h)$ and $q(h|x)$ to form complementary pairs of an {\em auto-encoder}
(since the input of one is the target of the other and vice-versa). Note that
from that point of view any of the two can act as encoder and the other as
decoder for it, depending on whether we start from $h$ or from $x$.
In the case of multiple latent layers,
each pair of conditionals $q(h^{(k+1)}|h^{(k)})$ and $p(h^{(k)} | h^{(k+1)})$
forms a symmetric auto-encoder, i.e., either one can act as the encoder
and the other as the corresponding decoder, since they are trained
with the same $(h^{(k)},h^{(k+1)})$ pairs (but with reversed roles of
input and target).

In addition, if noise is injected, e.g., in the form of the quantization
induced by a spike train,
then the trained auto-encoders are actually denoising auto-encoders, which
means that both the encoders and decoders are {\em contractive}: in the
neighborhood of the observed $(x,h)$ pairs, they map neighboring ``corrupted'' values
to the ``clean'' $(x,h)$ values.

\sb
\subsection{Joint Denoising Auto-Encoder with Latent Variables}
\sb

This suggests considering a special kind of ``joint'' denoising auto-encoder
which has the pair $(x,h)$ as ``visible'' variable, an auto-encoder that implicitly estimates 
an underlying $p(x,h)$. The transition operator\footnote{See Theorem 1
from~\citet{Bengio-et-al-NIPS2013-small} for the generative interpretation
of denoising auto-encoders: it basically states that one can sample from the
model implicitly estimated by a denoising auto-encoder by simply alternating
noise injection (corruption), encoding and decoding, these forming each step
of a generative Markov chain.} for that joint visible-latent
denoising auto-encoder is the following in the case of a single hidden layer:
\sa
\begin{align}
  (\tilde{x},\tilde{h}) \leftarrow {\rm corrupt}(x,h) \nonumber \\
  h \sim q(h | \tilde{x}) \;\;\;\;\;\;  x \sim p(x | \tilde{h}),
\end{align}
where the corruption may correspond to the stochastic quantization induced
by the neuron non-linearity and spiking process.
In the case of a middle layer $h^{(k)}$ in a deeper model, the transition operator
must account for the fact that $h^{(k)}$ can either be reconstructed from above or
from below, yielding, with probability say $\frac{1}{2}$,
\sa
\begin{align}
 h^{(k)} \sim p(h^{(k)} | \tilde{h}^{(k+1)}),
\sa
\end{align}
and with one minus that probability,
\sa
\begin{align}
 h^{(k)} \sim q(h^{(k)} | \tilde{h}^{(k-1)}).
\sa
\end{align}
\se
\sb

Since this interpretation provides a different model, it also provides
a different way of {\em generating} samples. Especially for shallow,
we have found that better samples could be obtained in this way, i.e., 
running the Markov chain with the above transition operator for a few steps.

There might be a geometric interpretation for the improved quality of the
samples when they are obtained in this way, compared to the directed
generative model that was defined earlier. Denote $q^*(x)$ the empirical
distribution of the data, which defines a joint
$q^*(h,x)=q^*(x)q^*(h|x)$. Consider the likely situation where $p(x,h)$ is not
well matched to $q^*(h,x)$ because for example the parametrization of $p(h)$
is not powerful enough to capture the complex structure in the empirical
distribution $q^*(h)$ obtained by mapping the training data through the
encoder and inference $q^*(h|x)$.  Typically, $q^*(x)$ would concentrate on a manifold and the
encoder would not be able to completely unfold it, so that $q^*(h)$ would
contain complicated structure with pockets or manifolds of high
probability. If $p(h)$ is a simple factorized model, then it will generate
values of $h$ that do not correspond well to those seen by the decoder
$p(x|h)$ when it was trained, and these out-of-manifold samples in
$h$-space are likely to be mapped to out-of-manifold samples in $x$-space.
One solution to this problem is to increase the capacity of $p(h)$
(e.g., by adding more layers on top of $h$). Another is to make $q(h|x)$
more powerful (which again can be achieved by increasing the depth of the
model, but this time by inserting additional layers below $h$). Now,
there is a cheap way of obtaining a very deep directed graphical model,
by unfolding the Markov chain of an MCMC-based generative model for
a fixed number of steps, i.e., considering each step of the Markov
chain as an extra ``layer'' in a deep directed generative model,
with shared parameters across these layers. As we have seen that there
is such an interpretation via the joint denoising auto-encoder over
both latent and visible, this idea can be immediately applied.
We know that each step of the Markov chain operator moves its
input distribution closer to the stationary distribution of the chain.
So if we start from samples from a very broad (say factorized) prior $p(h)$
and we iteratively encode/decode them (injecting noise appropriately as
during training) by successively sampling from $p(x|h)$ and then
from $q(h|x)$, the resulting $h$ samples should end up looking more
like those seen during training (i.e., from $q^*(h)$).

\sb
\subsection{Latent Variables as Corruption}
\sb

There is another interpretation of the training procedure, also as a denoising
auto-encoder, which has the advantage of producing a generative procedure
that is the same as the inference procedure except for $x$ being unclamped.

We return again to the generative interpretation of the denoising criterion
for auto-encoders, but this time we consider the non-parametric process
$q^*(h|x)$ as a kind of corruption of $x$ that yields the $h$ used as input
for reconstructing the observed $x$ via $p(x|h)$. Under that
interpretation, a valid generative procedure consists at each step in first
performing inference, i.e., sampling $h$ from $q^*(h|x)$, and second sampling
from $p(x|h)$. Iterating these steps generates $x$'s according to the
Markov chain whose stationary distribution is an estimator of the data
generating distribution that produced the training
$x$'s~\citep{Bengio-et-al-NIPS2013-small}. This view does not care about
how $q^*(h|x)$ is constructed, but it tells us that if $p(x|h)$ is trained to
maximize reconstruction probability, then we can sample in this way from
the implicitly estimated model.

We have also found good results using this procedure (Algorithm~\ref{alg:experiment-1} below), 
and from the point of
view of biological plausibility, it would make more sense that ``generating''
should involve the same operations as ``inference'', except for the input
being observed or not.

\sc
\section{Targetprop instead of Backprop}
\label{sec:bf-gradient}
\sb

In Algorithm~\ref{alg:variational-MAP} and the related stochastic variants
Eq.~\ref{eq:h-gradient} suggests that back-propagation
(through one layer) is still needed when $h^{(k)}$ is on the right hand side of the conditional
probability bar, e.g., to compute $\frac{\partial p(h^{(k-1}) | h^{(k)})}{\partial h^{(k)}}$.
Such a gradient is also the basic building block
in back-propagation for supervised learning: we need to back-prop through
one layer, e.g. 
to make $h^{(k)}$ more ``compatible'' with $h^{(k-1)}$. This provides
a kind {\em error signal}, which in the case of unsupervised learning
comes from the sensors, and in the case of supervised learning, comes
from the layer holding the observed ``target''.

Based on recent theoretical results on denoising 
auto-encoders, we propose 
the following estimator (up to a scaling constant) of the required gradient,
which is related to previous work on 
``target propagation''~\citep{Bengio-arxiv2014,Lee-et-al-arxiv2014-small} or 
targetprop for short.
To make notation simpler, we focus below on the case of two layers $h$ and $x$
with ``encoder'' $q(h|x)$ and ``decoder'' $p(x|h)$, and we want to estimate
$\frac{\partial \log p(x|h)}{\partial h}$. We start with the special case
where $p(x | h)$ is a Gaussian with mean $g(h)$ and $q(h | x)$ is Gaussian
with mean $f(x)$, i.e., $f$ and $g$ are the deterministic components 
of the encoder and decoder respectively.
The proposed estimator is then
\sb
\begin{equation}
\label{eq:bf-estimator}
 \widehat{\Delta h} = \frac{f(x) - f(g(h))}{\sigma^2_h},
\sb
\end{equation}
where $\sigma^2_h$ is the variance of the noise injected in $q(h|x)$. 

Let us now justify this estimator. Theorem 2 by~\citet{Alain+Bengio-ICLR2013-small}
states that in a denoising auto-encoder with reconstruction function $r(x)={\rm decode}({\rm encode}(x))$,
a well-trained auto-encoder estimates the log-score via the difference between
its reconstruction and its input:
\sb
\[
  \frac{r(x)-x}{\sigma^2} \rightarrow \frac{\partial \log p(x)}{\partial x},
\sb
\]
where $\sigma^2$ is the variance of injected noise, and $p(x)$ is the implicitly
estimated density. We are now going to consider two denoising auto-encoders
and apply this theorem to them. First, we note that the gradient 
$\frac{\partial \log p(x|h)}{\partial h}$ that we wish to estimate can be
decomposed as follows:
\sb
\[
  \frac{\partial \log p(x|h)}{\partial h} = \frac{\partial \log p(x,h)}{\partial h} -
    \frac{\partial \log p(h)}{\partial h}.
\]
Hence it is enough to estimate $\frac{\partial \log p(x,h)}{\partial h}$ as
well as $\frac{\partial \log p(h)}{\partial h}$.
The second one can be estimated by considering the auto-encoder which estimates $p(h)$ implicitly
and for which $g$ is
the encoder (with $g(h)$ the ``code'' for $h$) and $f$ is the decoder
(with $f(g(h))$ the ``reconstruction'' of $h$). Hence
we have that $\frac{f(g(h))-h}{\sigma_h^2}$ 
is an estimator of $\frac{\partial \log p(h)}{\partial h}$.

The other gradient can be estimated by considering the joint denoising
auto-encoder over $(x,h)$ introduced in the previous section. The 
(noise-free) reconstruction function for that auto-encoder is 
\sa
\[
  r(x,h) = (g(h),f(x)).
\sa
\]
Hence $\frac{f(x)-h}{\sigma_h^2}$ is an estimator of $\frac{\partial \log p(x,h)}{\partial h}$.
Combining the two estimators, we get
\sb
\[
  \frac{(f(x)-h)}{\sigma_h^2} - \frac{(f(g(h))-h)}{\sigma^2_h} = \frac{f(x)-f(g(h))}{\sigma^2_h},
\sa
\]
which corresponds to Eq.~\ref{eq:bf-estimator}.

\begin{figure}[ht]
\sa
\ifarxiv
\centerline{\includegraphics[width=.3\textwidth]{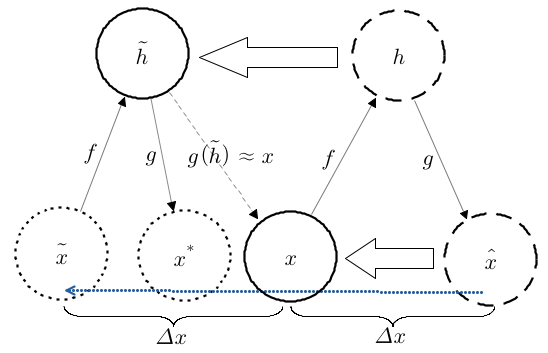}}
\else
\centerline{\includegraphics[width=.3\textwidth]{figures/back_step_BFAE.png}}
\fi
\sb
\caption{The optimal $h$ for maximizing $p(x|h)$ is $\tilde{h}$ s.t. $g(\tilde{h})=x$.
Since the encoder $f$ and decoder $g$ are approximate inverses of each other, their composition makes
a small move $\Delta x$. Eq.~\ref{eq:bf-estimator} is obtained by assuming that by considering an
$\tilde{x}$ at $x-\Delta$ and applying $f \circ g$, one would approximately recover $x$, which
should be true if the changes are small and the functions smooth (see ~\citet{Lee+Bengio-NIPSDL2014-small}
for a detailed derivation).
}
\label{fig:back-step}
\sc
\end{figure}

Another way to obtain the same formula from a geometric perspective
is illustrated in Figure~\ref{fig:back-step}. It
was introduced in ~\citet{Lee+Bengio-NIPSDL2014-small}
in the context of a backprop-free algorithm for training a denoising auto-encoder.

\begin{algorithm}[tb]
\ss
\caption{Inference, training and generation procedures used in Experiment 1.
The algorithm can naturally be extended to more layers.  $f_i()$ is the
feedforward map from layer $i-$1 to layer $i$ and $g_i()$ is the feedback map
from layer $i$ to layer $i-$1, with $x=h_0$ being layer 0.
}
\label{alg:experiment-1}
\begin{algorithmic}
\STATE Define INFERENCE($x$, $N$=15, $\delta$=0.1, $\alpha$=0.001):
\STATE Feedforward pass: $h_1 \leftarrow f_1(x)$, \hspace{5mm} $h_2 \leftarrow f_2(h_1)$
\FOR {$t=1$ to $N$}
  \STATE $h_2 \leftarrow h_2 + \delta (f_2(h_1) - f_2(g_2(h_2)))$ 
  \STATE $h_1 \leftarrow h_1 + \delta (f_1(x) - f_1(g_1(h_1))) + \alpha (g_2(h_2)-h_1)$ 
\ENDFOR
\STATE {\bf Return} $h_1$, $h_2$
\STATE
\STATE Define TRAIN() 
\FOR {$x$ in training set}
  \STATE $h_1, h_2 \leftarrow$ INFERENCE($x$) \; (i.e. E-part of EM)
  \STATE update each layer using local targets (M-part of EM)
  \STATE \hspace{0.5cm} $\Theta \leftarrow \Theta + \epsilon \frac{\partial}{\partial \Theta} \; (g_l(\tilde{h}_l) - h_{l-1})^2$
  \STATE \hspace{0.5cm} $\Theta \leftarrow \Theta + \epsilon \frac{\partial}{\partial \Theta} \; (f_l(\tilde{h}_{l-1}) - h_l)^2$
  \STATE where $\tilde{h}_l$ is a Gaussian-corrupted version of $h_l$.
  For the top sigmoid layer $\tilde{h}_2$ we average 3 samples
  from a Bernoulli distribution with $p(\tilde{h}_2=1)=h_2$ to obtain
  a {\it spike-like} corruption.
  %
\ENDFOR
\STATE Compute the mean and variance of $h_2$ using the training set. Multiply the variances by 4. Define $p(h_2)$ as sampling from this Gaussian.
\STATE
\STATE Define GENERATE():
\STATE Sample $h_2$ from $p(h_2)$
\STATE Assign $h_1 \leftarrow g_2(h_2)$ and $x \leftarrow g_1(h_1)$
\FOR {$t=1$ to $3$}
  \STATE $h_1,h_2 \leftarrow$ INFERENCE($x$,$\alpha=0.3$)
  \STATE $x \leftarrow g_1(h_1)$
\ENDFOR
\STATE {\bf Return} $x$
\end{algorithmic}
\ss
\end{algorithm}

\sc
\section{Related Work}
\label{sec:related-work}
\sb

An important inspiration for the proposed framework is the biological implementation
of back-propagation proposed by~\citet{Hinton-DL2007}. In that talk, Hinton
suggests that STDP corresponds to a gradient update step with the gradient on
the voltage potential corresponding to its temporal derivative. To obtain the supervised
back-propagation update in the proposed scenario would require symmetric weights
and synchronization of the computations in terms of feedforward and feedback
phases. 

Our proposal introduces a novel machine learning interpretation that also
matches the STDP behavior, based on a variational EM framework, allowing us
to obtain a more biologically plausible mechanism for deep generative unsupervised learning, 
avoiding the need for symmetric weights, and introducing a novel method to
obtain neural updates that approximately propagate gradients and move towards 
better overall configurations of neural activity (with difference target-prop).
There is also an interesting connection with an
earlier proposal for a more biologically plausible implementation of
supervised back-propagation~\citep{Xie+Seung-2003} which also relies on
iterative inference (a deterministic relaxation in that case), but
needs symmetric weights.

Another important inspiration is Predictive Sparse Decomposition (PSD)
~\citep{koray-psd-08}. PSD is a special case of
Algorithm~\ref{alg:variational-MAP} when there is only one layer and
the encoder $q(h|x)$, decoder $p(x|h)$, and prior $p(h)$ have a specific
form which makes $p(x,h)$ a sparse coding model and $q(h|x)$ a fast
parametric approximation of the correct posterior. Our proposal extends
PSD by providing a justification for the training criterion as a variational
bound, by generalizing to multiple layers of latent variables, and by
providing associated generative procedures.

The combination of a parametric approximate inference machine (the encoder)
and a generative decoder (each with possibly several layers of latent
variables) is an old theme that was started with the Wake-Sleep 
algorithm~\citep{Hinton95} and which finds very interesting instantiations in
the variational auto-encoder~\citep{Kingma+Welling-ICLR2014,Kingma-et-al-NIPS2014} and the reweighted wake-sleep
algorithm~\citep{Bornschein+Bengio-arxiv2014-small}. Two important differences
with the approach proposed here is that here we avoid back-propagation thanks to an inference step
that approximates the posterior. In this spirit, see the recent work introducing MCMC
inference for the variational auto-encoder~\citet{Salimans-et-al-arxiv2014}.

The proposal made here also owes a lot to the idea of target propagation
introduced in~\citet{Bengio-arxiv2014,Lee-et-al-arxiv2014-small}, to which it adds
the idea that in order to find a target that is consistent with both the input
and the final output target, it makes sense to perform iterative inference,
reconciling the bottom-up and top-down pressures. Addressing the weight
transport problem (the weight symmetry constraint) was also done for the
supervised case using ~{\em feedback alignment}~\citep{Lillicrap-et-al-arxiv2014}:
even if the feedback weights do not exactly match the feedforward weights,
the latter learn to align to the former and ``back-propagation'' (with the
wrong feedback weights) still works.

The targetprop formula avoiding back-propagation through one layer
is actually the same as proposed by~\citet{Lee+Bengio-NIPSDL2014-small} for
backprop-free auto-encoders. What has
been added here is a justification of this specific formula based on
the denoising auto-encoder theorem from~\citet{Alain+Bengio-ICLR2013-small},
and the empirical validation of its ability to climb the joint likelihood for
variational inference.

Compared to previous work on auto-encoders, and in particular their
generative interpretation~\citep{Bengio-et-al-NIPS2013-small,Bengio-et-al-ICML-2014}, this 
paper for the first time introduces latent variables without requiring 
back-propagation for training.

\section{Experimental Validation}




\begin{figure}[t]
\sa
\ifarxiv
\centerline{\includegraphics[width=.35 \textwidth]{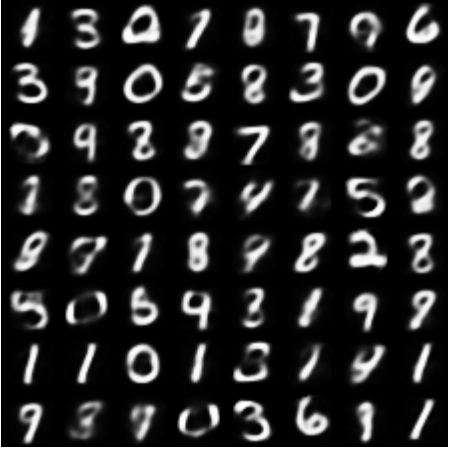}}
\else
\centerline{\includegraphics[width=.35 \textwidth]{figures/gen_samples.png}}
\fi
\sa
\caption{MNIST samples generated by GENERATE from 
Algorithm~\ref{alg:experiment-1} after training with TRAIN.}
\label{fig:samples-1}
\end{figure}

\begin{figure}[t]
\sa
\ifarxiv
\centerline{\includegraphics[width=.37 \textwidth]{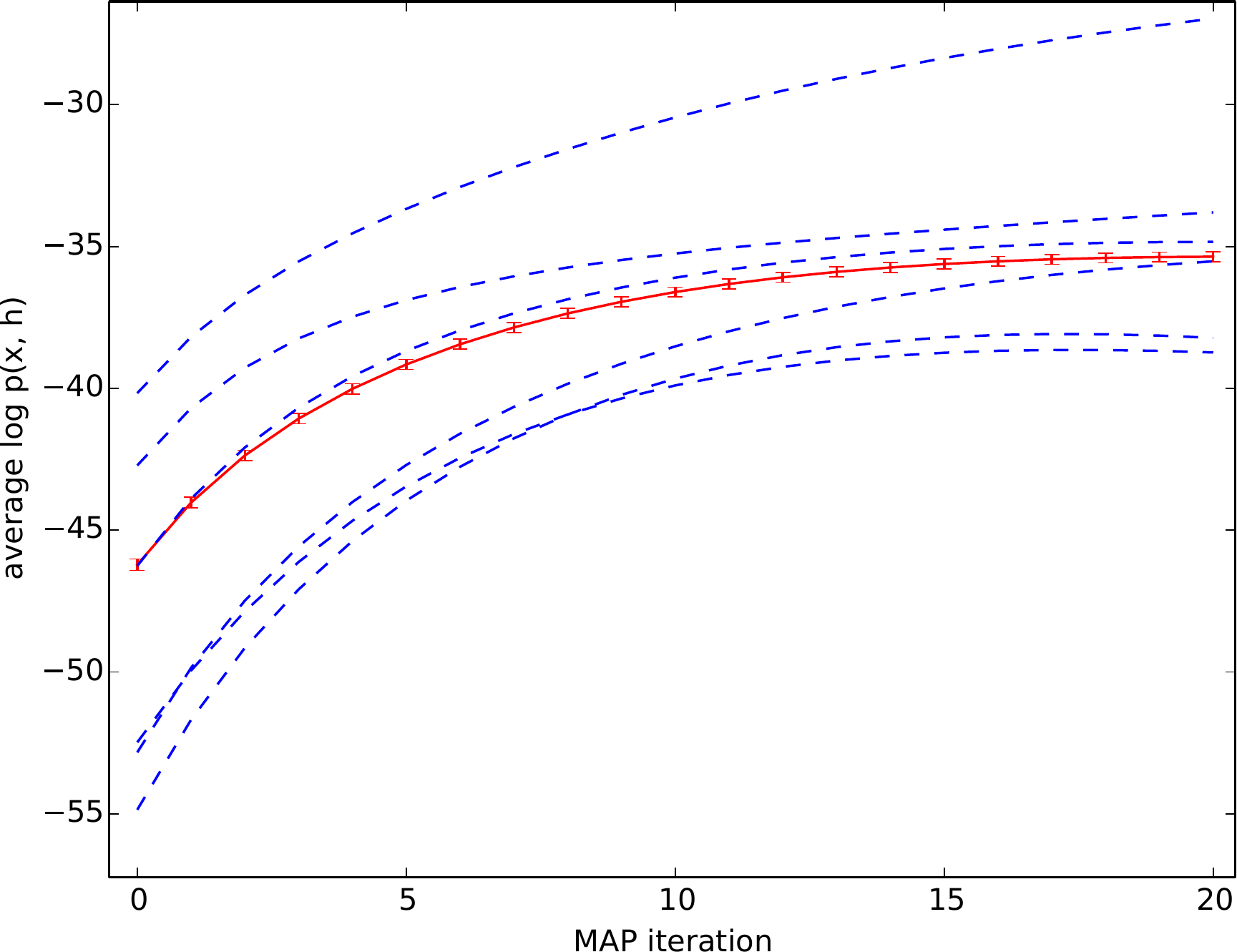}}
\else
\centerline{\includegraphics[width=.37 \textwidth]{figures/map-iterations.pdf}}
\fi
\sb
\caption{Increase of $\log p(x, h)$ over 20 iterations of the INFERENCE algorithm 
\ref{alg:experiment-1}, showing that the targetprop updates increase the
joint likelihood. The solid red line shows the average and the standard error 
over the full testset containing 10,000 digits. Dashed lines show $\log p(x,h)$ for individual 
datapoints.
}
\label{fig:map-iterations}
\end{figure}

\begin{figure*}[t]
\sb
\ifarxiv
\centerline{
\includegraphics[width=.30 \textwidth]{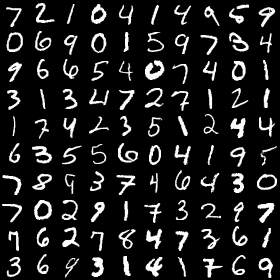}
\includegraphics[width=.30 \textwidth]{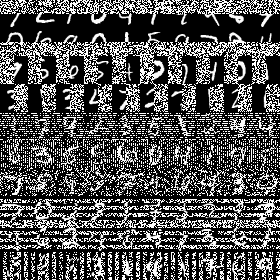}
\includegraphics[width=.30 \textwidth]{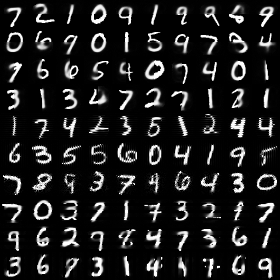}
}
\else
\includegraphics[width=.30 \textwidth]{figures/mnist-test-examples.png}
\includegraphics[width=.30 \textwidth]{figures/mnist-corrupted-examples.png}
\includegraphics[width=.30 \textwidth]{figures/mnist-inpainted-examples.png}
\fi
\sb
\caption{Examples of filling-in (in-painting) missing (initially corrupted)
parts of an image. Left: original MNIST test examples. Middle: initial
state of the inference, with half of the pixels randomly sampled (with
a different corruption pattern in each row of the figure). Right:
reconstructions using a variant of the INFERENCE procedure of Algorithm~\ref{alg:experiment-1}
for the case when some inputs are clamped.
}
\label{fig:inpainting}
\end{figure*}

Figure~\ref{fig:samples-1} shows generated samples obtained after training
on MNIST with Algorithm~\ref{alg:experiment-1} (derived from the
considerations of Sec.s~\ref{sec:deep}, \ref{sec:DAE-generator}
and~\ref{sec:bf-gradient}). The network has two hidden layers,
$h_1$ with 1000 softplus units and $h_2$ with 100 sigmoid units
(which can be considered biologically plausible~\citep{Glorot+al-AI-2011-small}).
We trained for 20 epochs, with minibatches of size 100 to speed-up
computation using GPUs. Results can be reproduced from code at {\small \tt http://goo.gl/hoQqR5}.
Using the Parzen density estimator 
previously used for that data, we obtain an estimated log-likelihood LL=236
(using a standard deviation of 0.2 for the Parzen density estimator, chosen
with the validation set), which
is about the same or better as was obtained for contractive
auto-encoders~\citep{Rifai+al-2011-small} (LL=121), deeper generative
stochastic networks~\citep{Bengio-et-al-ICML-2014} (LL=214) and generative
adversarial networks~\citep{Goodfellow-et-al-NIPS2014-small} (LL=225). In accordance with 
Algorithm~\ref{alg:experiment-1}, the variances of the
conditional densities are 1, and the top-level prior
is ignored during most of training (as if it was a very broad, uniform
prior) and only set to the Gaussian by the end of training, before
generation (by setting the parameters of $p(h_2)$ to the empirical mean and
variance of the projected training examples at the top level).
Figure~\ref{fig:map-iterations} shows that the targetprop updates
(instead of the gradient updates) allow the inference process to
indeed smoothly increase the joint likelihood. Note that if we
sample using the directed graphical model $p(x|h)p(h)$, the samples
are not as good and LL=126, suggesting as discussed in Sec.~\ref{sec:DAE-generator}
that additional inference and encode/decode iterations move $h$
towards values that are closer to $q^*(h)$ (the empirical distribution
of inferred states from training examples).
The experiment illustrated in Figure~\ref{fig:inpainting} shows that the
proposed inference mechanism can be used to fill-in missing values with a
trained model. The model is the same that was trained using
Algorithm~\ref{alg:experiment-1} (with samples shown in
Figure~\ref{fig:samples-1}). 20 iterations steps of encode/decode
as described below were performed, with a call to INFERENCE (to maximize
$p(x,h)$ over $h$) for each step, with a slight modification. Instead
of using $f_1(x)-f_1(g_1(h))$ to account for the pressure of $x$ upon
$h$ (towards maximizing $p(x|h)$), we used $f_1(x^v, g^m(h))-f(g(h))$,
where $x^v$ is the part of $x$ that is visible (clamped) while $g^m(h)$
is the part of the output of $g(h)$ that concerns the missing (corrupted)
inputs. This formula was derived from the same consideration as
for Eq.~\ref{eq:bf-estimator}, but where the quantity of interest
is $\frac{\partial \log p(x^v | h)}{\partial h}$ rather than
$\frac{\partial \log p(x | h)}{\partial h}$, and we consider that
the reconstruction of $h$, given $x^v$, fills-in the missing
inputs ($x^m$) from $g^m(h)$.

\sc
\section{Future Work and Conclusion}
\sb

We consider this paper as an exploratory step towards
explaining a central aspect of the brain's learning algorithm:
{\em credit assignment through many
  layers}. Of the non-plausible elements of back-propagation described
in the introduction, the proposed approach addresses all except the 5th.
As argued by~\citet{Bengio-arxiv2014,Lee-et-al-arxiv2014-small},
departing from back-propagation could be useful not just for biological
plausibility but from a machine learning point of view as well:
by working on the ``targets'' for the intermediate
layers, we may avoid the kind of reliance on smoothness and derivatives
that characterizes back-propagation, as these techniques can in principle work even
with highly non-linear transformations for which gradients are often near 0, e.g.,
with stochastic binary units~\citep{Lee-et-al-arxiv2014-small}.
Besides the connection between STDP and variational EM, an important
contribution of this paper is to show that the ``targetprop'' update
which estimates the gradient through one layer can be used for inference,
yielding systematic improvements in the joint likelihood and allowing to
learn a good generative model. Another interesting contribution is
that the variational EM updates, with noise added, can also be interpreted
as training a denoising auto-encoder over both visible and latent variables,
and that iterating from the associated Markov chain yields better samples
than those obtained from the directed graphical model estimated
by variational EM.

Many directions need to be investigated to follow-up on the work
reported here. An important element of neural circuitry is the
strong presence of lateral connections between nearby neurons
in the same area. In the proposed framework, an obvious place
for such lateral connections is to implement the prior on the
joint distribution between nearby neurons, something we have not
explored in our experiments. For example, ~\citet{GarriguesP2008}
have discussed neural implementations of the inference involved
in sparse coding based on the lateral connections.

Although we have found that ``injecting noise'' helped training a better
model, more theoretical work needs to be done to explore this replacement
of a MAP-based inference by an MCMC-like inference, which should help determine
how and how much of this noise should be injected.

Whereas this paper focused on unsupervised learning, these ideas could
be applied to supervised learning and reinforcement learning as well. 
For reinforcement learning, an important role of the proposed algorithms
is to learn to predict rewards, although a more challenging question is
how the MCMC part could be used to simulate future events. For both
supervised learning and reinforcement learning, we would probably want
to add a mechanism that would give more weight to minimizing prediction
(or reconstruction) error for some of the observed signals (e.g. $y$
is more important to predict than $x$).

Finally, a lot needs to be done to connect in more detail the proposals
made here with biology, including neural implementation 
using spikes with Poisson rates as the source of signal
quantization and randomness, taking into account the constraints on the
sign of the weights depending on whether the pre-synaptic neuron is
inhibitory or excitatory, etc.  In addition, although the operations
proposed here are backprop-free, they may still require some kinds of
synchronizations (or control mechanism) and specific connectivity to be
implemented in brains.

\ifarxiv
\section*{Acknowledgments} 
 
The authors would like to thank Jyri Kivinen, Tim Lillicrap and Saizheng Zhang
for feedback and discussions, as well as NSERC, CIFAR, Samsung
and Canada Research Chairs for funding, and Compute Canada
for computing resources.
\else
\fi


\bibliography{strings,ml}

\bibliographystyle{natbib}

\end{document}